\documentstyle[cogsci94]{article}

\newcommand{\text}[2]{\medskip\noindent\hangindent=55pt\hangafter=1
\hspace{15pt}\makebox[40pt][l]{\bf {#1}}\noindent{#2}\smallskip}

\input psfig
 
\title{\begin{small} In Proceedings of the Sixteenth Annual Conference of the Cognitive Science Society, pp 589-594, Lawrence Erlbaum Associates, 1994. \end{small} \\
Uniform Representations for Syntax-Semantics Arbitration
}
\author{
{\large \bf Kavi Mahesh} \and {\large \bf Kurt P. Eiselt} \\
College of Computing \\
Georgia Institute of Technology \\
Atlanta, GA 30332-0280 \\
{\tt \{mahesh,eiselt\}@cc.gatech.edu} \\
}

\begin{document}

\maketitle
 
\begin{abstract}
Psychological investigations have led to considerable insight into the
working of the human language comprehension system. In this article, we
look at a set of principles derived from psychological findings to argue
for a particular organization of linguistic knowledge along with a
particular processing strategy and present a computational model of
sentence processing based on those principles. Many studies have shown
that human sentence comprehension is an incremental and interactive
process in which semantic and other higher-level information interacts
with syntactic information to make informed commitments as early as
possible at a local ambiguity.  Early commitments may be made by using
top-down guidance from knowledge of different types, each of which must
be applicable independently of others.  Further evidence from studies of
error recovery and delayed decisions points toward an arbitration
mechanism for combining syntactic and semantic information in resolving
ambiguities. In order to account for all of the above, we propose that
all types of linguistic knowledge must be represented in a common form
but must be separable so that they can be applied independently of each
other and integrated at processing time by the arbitrator. We present
such a uniform representation and a computational model called COMPERE
based on the representation and the processing strategy.
\end{abstract}
 
\section{Introduction}
Psychological investigations of human language processing
have led to considerable insight into the working of the language
processor. Yet, attempts to build psychologically real computational
models of language processing face considerable problems in translating
the constraints put forth by psycholinguistic theories into the
representations and processes that constitute the computational model. 
In this article, we summarize a large body of
evidence from psycholinguistic literature into a set of principles and
use them to derive a computational model of human sentence processing
that employs a particular control of processing and a uniform
representation of all linguistic knowledge.
 
We argue that, in order to explain a large variety of findings in human
sentence comprehension, the computational model must employ a single
arbitration process that can integrate independently proposed syntactic
and semantic information to resolve ambiguities. We present a uniform
representation for syntactic and semantic knowledge that enables
syntactic and semantic interpretations to be integrated through
intermediate representations. The representational primitive is a node
that specifies part-of and has-part relations to other nodes,
preconditions on these relations, and expectations that could be generated
from such relations.
We show how such representations could be employed by a
parser that makes the right commitments at the appropriate times in
order to explain a variety of human sentence processing behaviors such
as incremental interpretation, immediacy of semantic and conceptual
interaction, modular behaviors such as purely structural preferences,
early commitment and error recovery, garden paths, and resource-limited
delayed decisions.  We have developed a model called COMPERE (Cognitive
Model of Parsing and Error Recovery) based on such an architecture and
tested it in a computer implementation.
 
\section{Psychological Constraints on  Language Comprehension Models}
Experimental observations of human sentence processing behavior can be 
summarized in terms of a set of principles including the following:
 
\subsection{1. Incremental Comprehension:} 
Psychological experiments have confirmed the general intuition that human
language comprehension is an incremental process
of progressively building a syntactic,
semantic, and referential interpretation of a sentence.
(e.g., Crain and Steedman, 1985; Marslen-Wilson and Tyler, 1987; Steedman, 1989; Taraban and McClelland, 1988). 
 
\subsection{2. Early Commitment:} 
The sentence processor must make early commitments as soon as it has the
information necessary to do so (e.g., Crain and Steedman, 1985; Frazier,
1987). Resource limitations such as working memory capacity and the
real-time nature of the comprehension task forbid the processor from
pursuing all possible paths in parallel until they are ruled out
(MacDonald, Just, and Carpenter, 1992). It must use information
available from every knowledge source to make a choice as early as
possible.
 
\subsection{3. Delayed Decisions:} However, the processor must
pursue parallel interpretations sometimes to account for delayed
decisions. When syntactic and semantic preferences are in conflict,
the processor must postpone the decision until further information
becomes available (Holmes et al., 1989) or until a decision is forced
by limits of memory resources.  Moreover, at the resource limit, it
appears that syntactic preferences override any semantic preferences
(Stowe, 1991).  Delayed decisions have also been observed in the
presence of lexical ambiguities.
 
\subsection{4. Error Recovery:}
Early commitment inevitably leads to erroneous decisions at times when
later information proves an earlier choice incorrect. The processor
must also be able to recover from such errors by switching to an
alternative interpretation (e.g., Eiselt, 1989; Stowe, 1991).  This
error recovery whether in structural or lexical ambiguity resolution,
should be possible through a repair process that is incremental.  Some
parts of the erroneous structures should be reused and others repaired
to result in correct structures.  Mere reprocessing from scratch does
not explain the grades of difficulty exhibited by weak and strong
garden-path sentences (Abney, 1989).
 
\subsection{5. Interaction:} Apart from incremental processing,
the processor supports interaction between different faculties such as
syntax, semantics, and reference (e.g., Crain and Steedman, 1985;
Marslen-Wilson and Tyler, 1987; Taraban and McClelland, 1988). Natural
languages are replete with ambiguities which cannot all be resolved by
the use of any one kind of knowledge. An incremental processor builds
interpretations of an incomplete sentence at all the different levels
that interact with each other and chooses an interpretation
that is best with respect to all the knowledge
available to the processor at that point. Incrementality, interaction,
early commitment, and error recovery can together explain the ability of
the human language processor to deal with the proliferation of local
ambiguities in natural languages (e.g., Crain and Steedman, 1985).
 
\subsection{6. Functional Independence:} 
A final constraint derivable from a vast body of psycholinguistic
studies can be summarized by stating that each kind of knowledge must
be capable of being applied independently of other kinds. For
instance, it must be possible to apply purely structural principles to
resolve an attachment ambiguity for an adjunct. This syntactic
knowledge must be represented so that it can be applied no matter what
lexical items are involved in the adjunct (Frazier, 1987; Frazier,
1989). Independence is also necessary to explain the behavior seen
when delays in commitment are terminated by resource limits (Stowe,
1991). Another important source of evidence for functional
independence between levels of language processing lies in behavioral
studies with aphasic subjects (e.g., Caramazza and Berndt, 1978).
Though the neurological independence between syntactic and semantic
knowledge has been questioned by recent studies of cross-linguistic
aphasia (Bates, Wulfeck, and MacWhinney, 1991), functional
independence would still be necessary to explain the differential
impairment of syntactic and semantic abilities in aphasia (Mahesh,
1993).

Moreover, in order to make sure that the actual course of steps in the
model agrees with human behavior, not just the end results of
processing, the representations, processes, and resources used by the
model must all be truthful to psychological data.
 
\section{The need to combine bottom-up and top-down processes}
Data-driven models of language comprehension with bottom-up strategies
are compatible with incremental semantic interpretation.  Incremental
comprehension is best described by a bottom-up strategy that interprets
each successively larger constituent as it is built from the next word
in the sentence.  A top-down strategy would force the processor to
commit to a whole constituent before analyzing the parts of the whole.
Making such commitments when the processor does not have necessary
information results in unwarranted backtracking.  
 
Standard theories of syntax using phrase-structure rules are
incompatible with incremental processing given that the steps taken by
the parser be psychologically real (Steedman, 1989).  Though it is
possible to apply a bottom-up strategy to phrase-structure rules, the
resulting process will not be incremental since the bottom-up parser
waits until it has seen every daughter of a constituent before
interpreting it.  Phrase-structure rules can be applied to carry out
an incremental interpretation only if sentences have a left-branching
structure which is not true with a majority of natural languages
(Steedman, 1989).  What we need is a bottom-up parser with top-down
guidance that can make early commitments before actually seeing every
part of a constituent so that semantic interpretation can be
incremental. Such early commitments may be made by employing top-down
influence from a variety of types of knowledge.
 
\subsection{Types of Top-Down Guidance to Bottom-Up Parsing}
Information providing top-down guidance to the
processor can be of three types: grammatical information about the
categories involved (Steedman, 1989), general structural principles
(Frazier, 1989), and feedback from semantics, reference and discourse
interpretation (e.g., Crain and Steedman, 1985; Taraban and McClelland,
1988).
 
\subsection{1. Syntactic Expectation:} 
Grammatical information tells the processor about the arguments that
must follow before the current constituent can be complete (and hence
grammatical). For instance, after seeing a noun phrase, a verb must
follow for the sentence to be complete. We can say that the processor
can {\em expect} to see a verb phrase at this point. The bottom-up
parser can use such syntactic expectations to make early commitments
at syntactic ambiguities.  This grammatical preference for expected
structures results in the same behavior as expected by the minimal
attachment principle and explains garden-path behavior in reduced
relative clause sentences such as (1).
 
\text{(1)}{The officers taught at the academy were very demanding.}
 
\subsection{2. Semantic and Pragmatic Preference:} 
One way to introduce extra-grammatical top-down influence on
attachment ambiguities is through interaction. Semantic and discourse
processes feedback to syntactic processing and exert preferences for
some attachments over others (e.g., Crain and Steedman, 1985; Stowe,
1991).  For instance, semantic feedback can tell the processor that
the prepositional phrase (PP) in sentence (2) must be attached to the
noun phrase (NP) since a horse cannot be used as an instrument for
seeing.
 
\text{(2)}{The officer saw the soldier with the horse.}
 
\subsection{3. Structural Preference:} 
The processor must also be able to exert purely structural preferences
that are independent of the categories and lexical items involved in
the ambiguous parts of the sentence. Examples of such a preference are
right association and minimal attachment (Frazier, 1987). Such
syntactic generalizations allow the syntactic processor to make early
commitments for adjuncts as well, thereby explaining several syntactic
phenomena (Frazier, 1989).  There is psychological evidence in
structural ambiguity resolution which demonstrates the need for this
kind of top-down influence. For instance, Stowe (1991) has shown that
the human sentence processor delays a decision (i.e., does not do an
early commitment) when there is a conflict between syntactic and
semantic preferences at a structural ambiguity (i.e., when top-down
influence from grammar or structural principles and those from
semantic feedback contradict each other).  Experiments showed that
people continue to delay the decision and pursue multiple
interpretations until they reach resource limits. At the limit, they
make a choice based only on syntactic preferences. This result would
be left unexplained but for the presence of a top-down influence of
the structural kind.
 
\subsection{The Control of Parsing}
Top-down guidance from syntactic expectations and structural preferences
can be integrated with a bottom-up control of parsing by employing a
form of left-corner parsing (Abney and Johnson, 1991).  We have
developed a variant of left-corner parsing by adding to it the virtues
of head-driven parsing. The resulting mechanism that we call {\em
Head-Signaled Left Corner Parsing} produces the right sequence of
syntactic commitments to account for a variety of data. The parser has
been implemented in the COMPERE model. Lack of space prevents us
from further describing the parsing mechanism here, but it is described
elsewhere (Mahesh, 1994).
 
\section{The Need for an Arbitration Mechanism}
While top-down and bottom-up influences on sentence comprehension
arising from syntactic sources of knowledge can be combined in the
parsing algorithm mentioned above, the sentence processor must also
combine information coming from semantic and conceptual analyses with
the syntactic preferences. This communication between syntax and
semantics is necessary to account for evidence from interaction,
delayed decisions, early commitment, and error recovery.  The
left-corner parsing algorithm merely identifies the points when the
communication ought to occur but doesn't tell us how the communication
is handled or how conflicts are resolved in the best interests of the
constraints on behavior.  One way to do this is to integrate the
representations of the two types of knowledge a priori as in semantic
grammars and what are called grammatical constructions in certain
other models (e.g., Jurafsky, 1991).  Such an approach suffers from
reduced generativity and other disadvantages (see the Related Work
section below).  In order to avoid losing generativity, the sentence
processor must keep the knowledge sources separate and introduce an
arbitration mechanism that dynamically combines information arising
from independent syntactic and semantic sources, resolving any
conflicts that might arise in the process.
 
The arbitrator needs to combine information coming from independent
syntactic and semantic sources which talk in different terms. Syntax
describes its interpretations in terms of grammatical relations such
as subject and object relations while semantics talks in terms of
thematic roles.\footnote{For purposes of illustration and simplicity,
we are employing thematic-role assignment using selectional
preferences from the lexicon as our theory of semantics. Our approach
however is not limited to thematic roles. For an example of a more
structured theory of semantics, see (Peterson and Billman, 1994).} The
arbitrator must establish correspondences between these two
representations (and between the decisions made by the two). One way
to bridge this communication gap between the languages of syntax and
semantics is to add a translation procedure in the arbitrator. This
procedure would translate grammatical relations to thematic relations
and vice versa using other kinds of linguistic information such as
active/passive voice, what roles go with a particular event, and so
on.
 
There are two problems with this approach. First, it is a procedural
solution with well-known disadvantages over a comparable declarative
solution.  Second, during  error recovery in incremental sentence
comprehension, corresponding things will have to be undone in syntax and
semantics. This is problematic if the only representation of the
correspondence  is in the arbitration procedures, since
correspondence knowledge is not present in the representations that
need to be manipulated by error recovery mechanisms.  Also, the only
kind of recovery possible is through reprocessing because there are no
 representations of intermediate decisions to repair or backtrack to.

We propose an alternative to this in which the gap is bridged via
intermediate representations. These intermediate roles connect
grammatical relations to thematic roles, resulting in a declarative
representation of the correspondence knowledge in the form of role
hierarchies.  Now, with this mechanism, during error recovery, the
processor can backtrack just the right amount and recover from the
error by repairing the erroneous structure rather than reprocessing
from scratch.

\begin{figure}[htbp]
\begin{center}
\ \psfig{figure=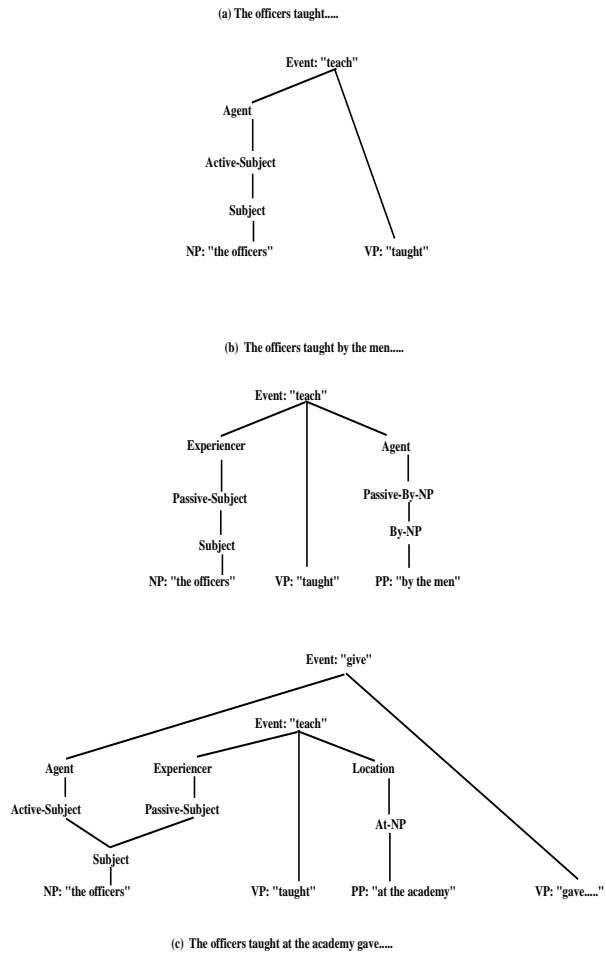,width=3.25in,height=5.0in}
\caption{Role Hierarchies: From Phrases to Thematic Roles.}
\end{center}
\end{figure}

Role hierarchies that bridge the gap between syntactic phrases and
thematic roles for example sentences are shown in Figure
1.\footnote{For simplicity, syntactic structures that are linked to
the role hierarchies are not shown.} As the sentence progresses from
that in (a) to the one in (b) or the one in (c), we can see the role
structures being repaired to recover from local errors. It can also be
seen that while both (b) and (c) involve repairs to role hierarchies,
(c) also involves a reorganization of its syntactic structure to
accommodate its reduced relative structure. Thus (c) is more of a
garden-path sentence than (b).

\section{The Uniform Representation}
In order for the single arbitration process to manipulate both
syntactic and semantic types of knowledge both during early
commitments and error recovery, we propose a uniform representation of
all types of knowledge.  The elements of the uniform representation
are units called {\em nodes} comprised of (a) part-whole links, (b)
preconditions, (c) expectations, and (d) preference levels. The
representations are to be read as (Fig. 2a) ``the parts can be linked
to the wholes when the preconditions are met and if so, the
corresponding expectations can be generated at that point.''  A node
represents all the knowledge about a syntactic or semantic category.
Sample representations of syntactic and semantic knowledge in this
general form are shown in parts (b) and (c) of Figure 2.  The NP node
specifies the wholes S, VP, and PP of which it can be a part along
with the preconditions and expectations therein. Similarly, the
semantic node specifies knowledge of the thematic roles that go with a
{\em Teach} event and the preconditions on the role fillers.  The
examples shown in Figure 2 are, however, simplified and incomplete.

Intermediate roles such as Active-Subject, Non-Agent-Active-Subject, and
Passive-By-NP are used to bridge the grammatical and thematic relations
between the parts of a sentence. Their use in backtracking during error
recovery can be seen by examining the transition from (a) to (b) and (a)
to (c) in the examples in Figure 1.
 
\begin{figure}[htbp]
\begin{center}
\ \psfig{figure=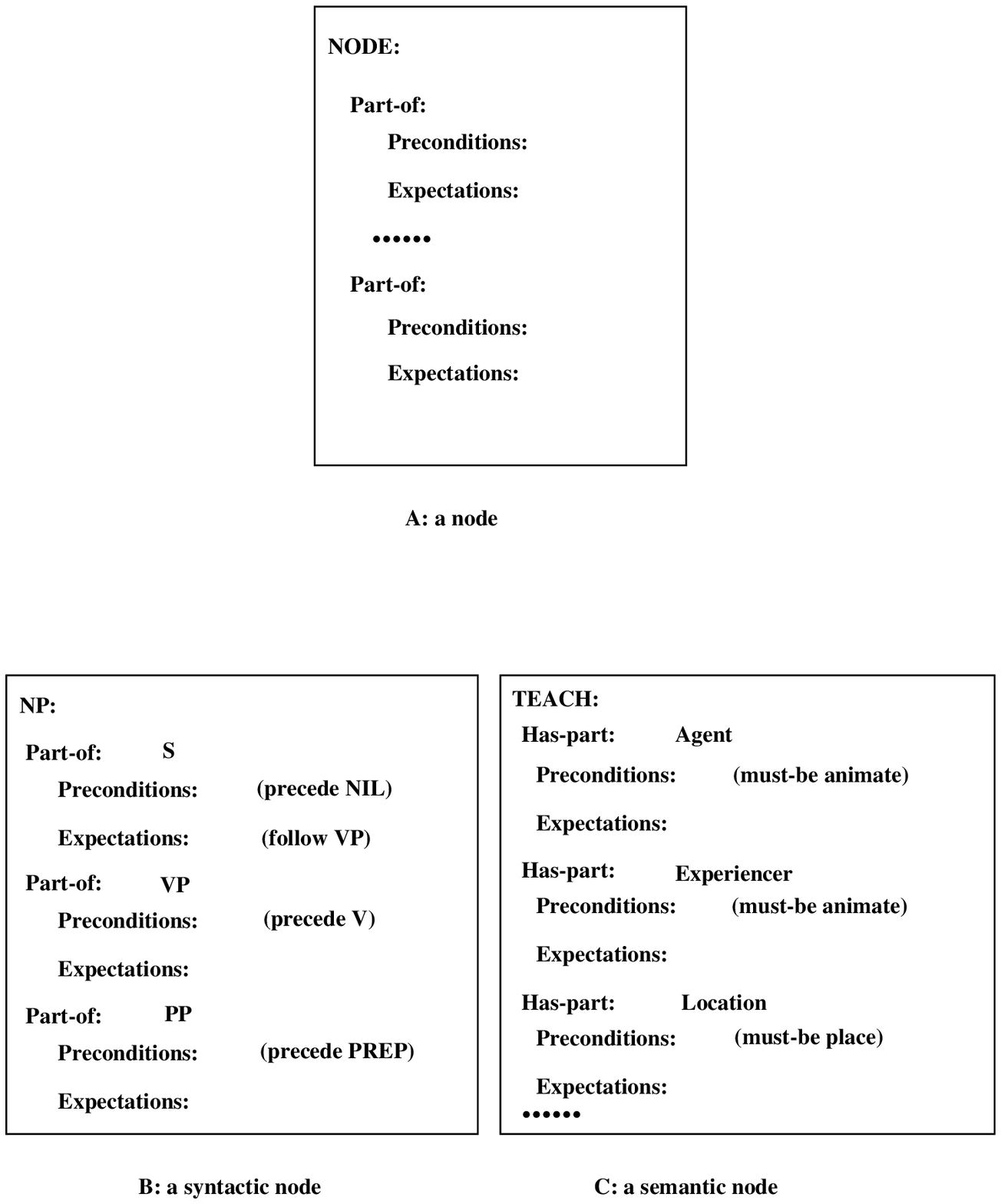,width=3.25in,height=4.0in}
\caption{Uniform Representation of Linguistic Knowledge.}
\end{center}
\end{figure}

As an alternative to uniform representations, if there are separate
representations of different types of knowledge in different forms,
it becomes hard to support incremental interaction in view of the constraint
that the time course of actions taken by the processor must be
psychologically real. Especially in the case of error recovery
phenomena, it is elegant to have an arbitration process which processes
all the kinds of knowledge (i.e., applies common operations on different
kinds of knowledge) so that it can make corresponding changes to
syntactic and semantic interpretations while recovering from an error.

\subsection{Processing with Uniform Representations}
The arbitration process has been implemented in a sentence understanding
program called COMPERE. Its single arbitration process reads words in a
left-to-right order, proposes ways of attaching them syntactically and
semantically to previously built structures, follows the left-corner
parsing algorithm mentioned above in producing the right sequence of
decisions, and selects the  most preferred attachments at each point
by combining the preferences assigned to the alternative attachments
proposed bottom-up with the grammatical, semantic, and structural types
of top-down influence.  COMPERE produces tree structures representing
the syntactic structure and thematic role assignment for a sentence.
It is capable of representing and pursuing multiple interpretations in
parallel when a conflict between syntactic and semantic preferences
forces a delay in resolving an ambiguity. Further details of COMPERE's
processing mechanisms can be found in (Mahesh, 1993; Eiselt et al,
1993).

COMPERE's arbitration process retains alternative interpretations that
it did not select in the first place.\footnote{For an account of the
Conditional Retention theory of error recovery, see (Eiselt 1989).}
If an error is detected at a later point (when there is no way to
attach a new constituent, for instance), it recovers from the error by
switching over to one of the retained alternatives, performing any
repairs to the syntactic and role structures (as shown in Fig. 1 for
example) (Eiselt et al., 1993; Holbrook et al., 1992). The
intermediate roles in the role hierarchies help maintain the
correspondence between the changes made to the syntactic and semantic
structures. COMPERE maintains the correspondence information by
uniformly connecting every syntactic node to its corresponding
semantic node(s) (including the intermediate roles) and vice versa.

In order to model the variety of behaviors, including the findings of
Stowe (1991) on limited delayed decisions, the arbitration process must be
enhanced by imposing architectural constraints, such as resource limits.
The control structure of the process can then order the preferences
based on their cost in terms of resources such as working memory. For
instance, limited delayed decisions can be modeled by ordering semantic
influence on par with syntactic preferences  to begin with and by ignoring the
semantic preferences when the processor has run out of resources.
COMPERE's separation of knowledge sources retains the functional
independence between them which is necessary for the above enhancements.
 
COMPERE can deal with a variety of (psycho)linguistically interesting
constructs such as relative clause sentences, complements, and
prepositional adjuncts with both lexical and structural ambiguities
(including the ones used in the examples above). It can analyze their
syntax and assign thematic roles. For example, it can show why sentence
(1) is a garden-path sentence using grammatical expectations and early
commitment; it can recover from the garden path to reinterpret the
sentence; and it can use immediate semantic interaction to show why
sentence (3) is not a garden path.  The program has demonstrated the
computational feasibility of the uniform representation and the
arbitration process. However, its semantic competence is limited to
thematic role assignment and at this time it does not have the knowledge
to carry out discourse and reference processing.
 
\text{(3)}{The courses taught at the academy were very demanding.}

\section{Related Work}
Though incremental interaction between syntax and semantics is not a
new idea in computational modeling of human sentence comprehension,
previous models that subscribed to such an interactive view have
sacrificed the ability to apply syntactic and semantic knowledge
independently of each other. While syntax-semantics interaction helps
us explain a variety of psychological data, it is certainly not
sufficient to account for the data on purely structural preferences,
on deferred decisions with limited delays, and on error recovery
phenomena (Eiselt et al., 1993).  Below we briefly describe why some
of the other models cannot explain the apparently incompatible data on
modular effects (which functional independence can explain) as well as
interactive effects (which incremental interaction can explain).

\noindent
{\bf Models with Integrated Representations:} These models resort to
an {\em a priori} integration of knowledge sources with a consequent
loss of functional independence (e.g., Jurafsky, 1991).

\noindent
{\bf Categorial Grammars:} Categorial grammars which establish a
strong correspondence between syntactic and semantic categories
(Steedman, 1989) also lack functional independence between syntax and
semantics.

\noindent
{\bf NL-SOAR:} Though this model, a contemporary of COMPERE, has a
uniform representation and integrates multiple knowledge sources to
resolve ambiguities, it cannot account for delayed decisions since it
can only maintain one interpretation at a time. In addition, its
chunking operations result in the integration of different types of
knowledge with a consequent loss in functional independence (e.g.,
Lehman et al., 1991; Lewis, 1993).

\noindent
{\bf Connectionist Models:} Connectionist models of sentence
processing (e.g., Waltz and Pollack, 1985), including the Competition
Model of MacWhinney and Bates (Bates et al., 1991), also have uniform
representations.  However, they do not deal very well with the full
syntactic complexity of natural language.  It is yet to be seen if the
simple computational mechanisms of activation and inhibition in a
network can exercise enough control of processing to model the precise
mechanisms of arbitration, error recovery, and delayed decisions.
 
\section{Conclusion}
Table \ref{conclusion-table} summarizes the features of COMPERE that
help us satisfy the psychological constraints we started with. \\

\begin{table}
\caption{How COMPERE meets psychological constraints.}
\label{conclusion-table}
\begin{small}
\begin{center}
\begin{tabular}{l l}
\hline
Constraint & Features of COMPERE \\
\hline
1. Incrementality & Head-Signaled Left-Corner \\
    & parsing algorithm. \\
2. Early Commitment & Grammatical, structural, and \\
   & semantic top-down guidance \\
  & to the parser. \\
3. Delayed Decisions & Arbitration mechanism and \\
  & retention of alternatives. \\
4. Error Recovery &  Arbitration mechanism, \\
   & retention, uniform rep., \\
	& declarative rep. of syntax-\\
         & semantics correspondence \\
	& through intermediate roles. \\
5. Semantic Interaction & Arbitration of independently \\
 & proposed syntactic and \\
 &  semantic preferences. \\
6. Functional Independence & Separate representation of \\
   & syntactic and semantic \\
   & knowledge. \\
\hline
\end{tabular}
\end{center}
\end{small}
\end{table}

\noindent
We have shown that in order to explain a variety of human sentence
processing behaviors, the sentence processor must use a bottom-up
strategy and yet accommodate top-down influence from grammatical,
semantic, discourse, and structural preferences. We presented a uniform
representation of different kinds of knowledge in a common format but in
separate units.  Using this representation, we showed how we can arbitrate
syntactic and semantic processes and account for a range of behaviors.
We have demonstrated the computational feasibility of the model in the
COMPERE program. We believe that the combination of an arbitration mechanism
and a uniform representation will take us a long way in modeling human
sentence processing behavior. \\
 
\section{References}

\begin{description}\setlength{\labelsep}{0pt}

\item [] S. P. Abney (1989).
 A Computational Model of Human Parsing.
   {\em Journal of Psycholinguistic Research,} 18(1):129--144.

\item [] S. P. Abney and M. Johnson (1991).
Memory Requirements and Local Ambiguities of Parsing 
Strategies. {\em Journal of Psycholinguistic Research,} 20(3):233-250. 

\item [] E. Bates, B. Wulfeck, and B. MacWhinney (1991). Cross-Linguistic 
Research in Aphasia: An Overview.
{\em Brain and Language,} 41(2):123-148.

\item [] A. Caramazza and R. S. Berndt (1978). Semantic and syntactic 
processes in aphasia: A review
of the literature. {\em Psychological Bulletin,} 85:898-918.

\item [] S. Crain and M. Steedman (1985).
 On not being led up the garden path: the use of context by the
  psychological syntax processor.
 In D.R. Dowty, L. Karttunen, and A.M. Zwicky, editors,   {\em Natural
  Language Parsing: Psychological, computational, and theoretical
  perspectives.} Cambridge University Press.

\item [] K. P. Eiselt (1989).
   {\em Inference Processing and Error Recovery in Sentence
  Understanding.}
 Ph.D. thesis, University of California, Irvine, CA.
 Tech. Report 89-24.

\item [] K. P. Eiselt, K. Mahesh, and J. K. Holbrook (1993).
Having Your Cake and Eating It Too: Autonomy
and Interaction in a Model of Sentence Processing. In {\em Proceedings of the 
Eleventh National Conference on AI} (AAAI-93), pp 380-385.

\item [] L. Frazier (1987).
 Theories of Sentence Processing.
 In J.L. Garfield, editor,   {\em Modularity in Knowledge
  Representation and Natural Language Understanding.} MIT Press.

\item [] L. Frazier (1989).
 Against Lexical Generation of Syntax.
 In W. Marslen-Wilson, editor,   {\em Lexical Representation and
  Process.} MIT Press.

\item [] J. K. Holbrook, K. P. Eiselt, and K. Mahesh (1992).
 A Unified Process Model of Syntactic and Semantic Error Recovery in
  Sentence Understanding.
 In   {\em Proceedings of the Fourteenth Annual Conference of the
  Cognitive Science Society,} pages 195--200.

\item [] V. M. Holmes, L. Stowe, and L. Cupples (1989). Lexical Expectations 
in Parsing 
Complement-Verb Sentences. {\em Journal of Memory and Language,} 28:668-689.

\item [] D. Jurafsky (1991).
 An On-Line Model of Human Sentence Interpretation.
 In   {\em Proceedings of the Thirteenth Annual Conference of the
  Cognitive Science Society,} pages 449--454.

\item [] J. F. Lehman, R. L. Lewis, and A. Newell (1991).
 Integrating Knowledge Sources in Language Comprehension.
 In   {\em Proceedings of the Thirteenth Annual Conference of the
  Cognitive Science Society,} pages 461--466.

\item [] R. L. Lewis (1993). An Architecturally Based Theory of Human 
Sentence Comprehension.  PhD Thesis, Computer Science Department,
Carnegie Mellon University, Pittsburgh, PA.  Technical Report
CMU-CS-93-226.

\item [] M. C. MacDonald, M. A. Just, and P. A. Carpenter (1992).
 Working Memory Constraints on the Processing of Syntactic Ambiguity.
{\em Cognitive Psychology,} 24:56--98.

\item [] K. Mahesh (1993). A Theory of Interaction and Independence in 
Sentence Understanding.  Technical Report GIT-CC-93/34, College of
Computing, Georgia Institute of Technology, Atlanta, GA.

\item [] K. Mahesh (1994). Reaping the benefits of interactive syntax and 
semantics. To appear in the {\em Proceedings of the 32nd Annual
Meeting of the Association for Computational Linguistics,} 27 June - 1
July 1994.

\item [] W. Marslen-Wilson and L. K. Tyler (1987).
 Against Modularity.
 In J.L. Garfield, editor,   {\em Modularity in Knowledge
  Representation and Natural-Language Understanding.} MIT Press.

\item [] J. Peterson and D. Billman (1994). 
Correspondences between Syntactic Form and Meaning: From Anarchy to
Hierarchy.  To appear in the {\em Proceedings of the Sixteenth Annual
Conference of the Cognitive Science Society,} Aug 13-16, 1994.

\item [] M. J. Steedman (1989).
 Grammar, Interpretation, and Processing from the Lexicon.
 In W. Marslen-Wilson, editor,   {\em Lexical Representation and
  Process.} MIT Press.

\item [] L. A. Stowe (1991).
 Ambiguity Resolution: Behavioral Evidence for a Delay.
 In  {\em Proceedings of the Thirteenth Annual Conference of the
  Cognitive Science Society,} pages 257--262.

\item [] R. Taraban and J. L. McClelland (1988).
 Constituent Attachment and Thematic Role Assignment in Sentence
  Processing: Influences of Content-Based Expectations.
   {\em Journal of Memory and Language,} 27:597--632.

\item [] D. L. Waltz and J. B. Pollack (1985).
 Massively Parallel Parsing: A Strongly Interactive Model of Natural
  Language Interpretation.
   {\em Cognitive Science,} 9:51--74.

\end{description}

\end{document}